# Causes and Explanations: A Structural-Model Approach—
## Part I: Causes

**Joseph Y. Halpern**
Cornell University
Dept. of Computer Science
Ithaca, NY 14853
halpern@cs.cornell.edu
http://www.cs.cornell.edu/home/halpern

**Judea Pearl**
Dept. of Computer Science
University of California, Los Angeles
Los Angeles, CA 90095
judea@cs.ucla.edu
http://www.cs.ucla.edu/~judea

## Abstract

We propose a new definition of *actual causes*, using *structural equations* to model counterfactuals. We show that the definition yields a plausible and elegant account of causation that handles well examples which have caused problems for other definitions and resolves major difficulties in the traditional account.

## 1 INTRODUCTION

What does it mean that an event $A$ *actually causes* event $B$? This is a question that goes beyond mere philosophical speculation. As Good [1993] argues persuasively, in many legal settings, what needs to be established (for determining responsibility) is not a counterfactual kind of causation, but "cause in fact." A typical example considers two fires advancing toward a house. If fire $A$ burned the house before fire $B$, we (and many juries nationwide) would consider fire $A$ "the actual cause" for the damage, even supposing the house would have definitely burned down by fire $B$, if it were not for $A$. Actual causation is also important in artificial intelligence applications. Whenever we undertake to *explain* a set of events that unfold in a specific scenario, the explanation produced must acknowledge the actual cause of those events. The automatic generation of adequate explanations, a task essential in planning, diagnosis and natural language processing, therefore requires a formal analysis of the concept of actual cause.

Giving a precise and useful definition of actual causality is notoriously difficult. The philosophical literature has been struggling with this notion since the days of Hume [1739]. (See [Sosa and Tooley 1993], [Hall 2002], and [Pearl 2000] for some recent discussions.) To borrow just one example from Hall [2002], suppose a bolt lightning hits a tree and starts a forest fire. It seems reasonable to say that the lightning bolt is a cause of the fire. (Indeed, the description "the lightning bolt ... starts a forest fire" can be viewed as saying this.) But what about the oxygen in the air and the fact that the wood was dry? Presumably, if there has not been oxygen or the wood was wet there would not have been a fire. Carrying this perhaps to the point of absurdity, what about the Big Bang? This problem is relatively easy to deal with, but there are a host of other, far more subtle, difficulties that have been raised over the years.

Here we give a definition of actual causality based on the language of *structural equations*; in a companion paper ([Halpern and Pearl 2001]; see also the full paper [Halpern and Pearl 2000]), we give a definition of *(causal) explanation* using the definition of causality. The use of structural equations as a model for causal relationships is standard in the social sciences, and seems to go back to the work of Sewall Wright in the 1920s (see [Goldberger 1972] for a discussion); the framework we use here is due to Pearl [1995], and is further developed in [Galles and Pearl 1997; Halpern 2000; Pearl 2000]. While it is hard to argue that our definition (or any other definition, for that matter) is the "right definition", we show that it deals well with the difficulties that have plagued other approaches in the past, especially those exemplified by the rather extensive compendium of Hall [2002].

There has been extensive discussion about causality in the literature, particularly in the philosophy literature. To keep this paper to manageable length, we spend only minimal time describing other approaches and comparing ours to them. We refer the reader to [Hall 2002; Pearl 2000; Sosa and Tooley 1993; Spirtes, Glymour, and Scheines 1993] for details and criticism of the probabilistic and logical approaches to causality in the philosophy literature. (We do try to point out where our definition does better than perhaps the best known approach, due to Lewis [1986, 2000] in the course of discussing the examples.)

There has also been work in the AI literature on causality. Perhaps the closest to this are papers by Pearl and his colleagues that use the structural-model approach. The definition of causality in this paper was inspired by an earlier paper of Pearl's [1998] that defined actual causality in terms of a construction called a *causal beam*. The causal beam definition was later modified somewhat (see [Pearl 2000, Chapter 10]), largely due to the considerations addressed in this paper. The definition given here is more transparent and handles a number of cases better (see Example 4.4).

Tian and Pearl [2000] give results on calculating the probability that $A$ is a *necessary* cause of $B$—that is, the probability that $B$ would not have occurred if $A$ had not occurred.

Necessary causality is related to but different from actual causality, as the definitions should make clear. Other work (for example, [Heckerman and Shachter 1995]) focuses on when a random variable $X$ is the cause of a random variable $Y$; by way of contrast, we focus on when an *event* such as $X = x$ causes an event such as $Y = y$. As we shall see, many of the subtleties that arise when dealing with events simply disappear if we look at causality at the level of random variables. Finally, there is also a great deal of work in AI on formal action theory (see, for example, [Lin 1995; Sandewall 1994]), which is concerned with the proper way of incorporating causal relationships into a knowledge base so as to guide actions. The focus of our work is quite different; we are concerned with extracting the actual causality relation from such a knowledge base, coupled with a specific scenario.

## 2   CAUSAL MODELS: A REVIEW

We briefly review the basic definitions of causal models, as defined in terms of structural equations, and the syntax and semantics of a language for reasoning about causality. See [Galles and Pearl 1997; Halpern 2000; Pearl 2000] for more details, motivation, and intuition.

**Causal Models:** The basic picture here is that the world is described by random variables, some of which may have a causal influence on others. This influence is modeled by a set of *structural equations*, where each equation represents a distinct mechanism (or law) in the world, one that may be modified (by external actions) without altering the others. In practice, it seems useful to split the random variables into two sets, the *exogenous* variables, whose values are determined by factors outside the model, and the *endogenous* variables. It is these endogenous variables whose values are described by the structural equations.

More formally, a *signature* $S$ is a tuple $(\mathcal{U}, \mathcal{V}, \mathcal{R})$, where $\mathcal{U}$ is a finite set of exogenous variables, $\mathcal{V}$ is a finite set of endogenous variables, and $\mathcal{R}$ associates with every variable $Y \in \mathcal{U} \cup \mathcal{V}$ a nonempty set $\mathcal{R}(Y)$ of possible values for $Y$ (that is, the set of values over which $Y$ ranges). A *causal model* (or *structural model*) over signature $S$ is a tuple $M = (S, \mathcal{F})$, where $\mathcal{F}$ associates with each variable $X \in \mathcal{V}$ a function denoted $F_X$ such that $F_X : (\times_{U \in \mathcal{U}} \mathcal{R}(U)) \times (\times_{Y \in \mathcal{V} - \{X\}} \mathcal{R}(Y)) \to \mathcal{R}(X)$. $F_X$ tells us the value of $X$ given the values of all the other variables in $\mathcal{U} \cup \mathcal{V}$.

**Example 2.1:** Suppose that we want to reason about a forest fire that could be caused by either lightning or a match lit by an arsonist. Then the causal model would have the following endogenous variables (and perhaps others):

- $F$ for fire ($F = 1$ if there is one, $F = 0$ otherwise)
- $L$ for lightning ($L = 1$ if lightning occurred, $L = 0$ otherwise)
- $ML$ for match lit ($ML = 1$ if the match was lit and $ML = 0$ otherwise).

The set $\mathcal{U}$ of exogenous variables includes things we need to assume so as to render all relationships deterministic (such as whether the wood is dry, there is enough oxygen in the air, etc.). If $\vec{u}$ is a setting of the exogenous variables that makes a forest fire possible (i.e., the wood is sufficiently dry, there is oxygen in the air, and so on) then, for example, $F_F(\vec{u}, L, ML)$ is such that $F = 1$ if $L = 1$ or $ML = 1$. ∎

Given a causal model $M = (S, \mathcal{F})$, a (possibly empty) vector $\vec{X}$ of variables in $\mathcal{V}$, and vectors $\vec{x}$ and $\vec{u}$ of values for the variables in $\vec{X}$ and $\mathcal{U}$, we can define a new causal model denoted $M_{\vec{X} \leftarrow \vec{x}}$ over the signature $S_{\vec{X}} = (\mathcal{U}, \mathcal{V} - \vec{X}, \mathcal{R}|_{\mathcal{V} - \vec{X}})$. Intuitively, this is the causal model that results when the variables in $\vec{X}$ are set to $\vec{x}$ by some external action that affects only the variables in $\vec{X}$; we do not model the action or its causes explicitly. Formally, $M_{\vec{X} \leftarrow \vec{x}} = (S_{\vec{X}}, \mathcal{F}^{\vec{X} \leftarrow \vec{x}})$, where $F_Y^{\vec{X} \leftarrow \vec{x}}$ is obtained from $F_Y$ by setting the values of the variables in $\vec{X}$ to $\vec{x}$.

It may seem strange that we are trying to understand causality using causal models, which clearly already encode causal relationships. Our reasoning is not circular. Our aim is not to reduce causation to noncausal concepts, but to interpret questions about causes of specific events in fully specified scenarios in terms of generic causal knowledge such as what we obtain from the equations of physics. The causal models encode background knowledge about the tendency of certain event types to cause other event types (such as the fact that lightning can cause forest fires). We use the models to determine the causes and explanations of single events, such as whether it was arson that caused the fire of June 10, 2000, given what is known or assumed about that particular fire.

We can describe (some salient features of) a causal model $M$ using a *causal network*. This is a graph with nodes corresponding to the random variables in $\mathcal{V}$ and an edge from a node labeled $X$ to one labeled $Y$ if $F_Y$ depends on the value of $X$. Intuitively, variables can have a causal effect only on their descendants in the causal network; if $Y$ is not a descendant of $X$, then a change in the value of $X$ has no affect on the value of $Y$. In this paper, we restrict attention to what are called *recursive* (or *acyclic*) equations; these are ones that can be described with a causal network that is a dag. It should be clear that if $M$ is a recursive causal model, then there is always a unique solution to the equations in $M_{\vec{X} \leftarrow \vec{x}}$, given a setting $\vec{u}$ for the variables in $\mathcal{U}$ (we call such a setting $\vec{u}$ a *context*).

As we shall see, there are many nontrivial decisions to be made when choosing the structural model. The exogenous variables to some extent encode the background situation, that which we wish to take for granted. Other implicit background assumptions are encoded in the structural equations themselves. Suppose that we are trying to decide whether a lightning bolt or a match was the cause of the forest fire, and we want to take for granted that there is sufficient oxygen in the air and the wood is dry. We could model the dryness of the wood by an exogenous variable $D$ with values 0 (the wood is wet) and 1 (the wood is dry). By making $D$ exogenous, its value is assumed to be given and out of the control of the modeler. We could also take the amount of oxygen as an exogenous variable (for example, there could be a variable $O$ with two values—0, for insufficient oxygen, and 1, for sufficient oxygen); alternatively, we could choose not to model oxygen explicitly at all. For example, suppose we have, as before, a random variable $ML$ for match lit, and another variable $WB$ for wood burning, with values 0 (it's not) and 1 (it is). The structural equation $F_{WB}$ would describe the dependence of $WB$ on $D$ and $ML$. By setting $F_{WB}(1,1) = 1$, we are saying that the wood will burn if the match is lit and the wood is dry. Thus, the equation is implicitly modeling our assumption that there is sufficient oxygen for the wood to burn. If we were to explicitly model the amount of oxygen in the air (which certainly might be relevant if we were analyzing fires on Mount Everest), then $F_{WB}$ would also take values of $O$ as an argument.

Besides encoding some of our implicit assumptions, the structural equations can be viewed as encoding the causal mechanisms at work. Changing the underlying causal mechanism can affect what counts as a cause. Section 4 provides several examples of the importance of the choice of random variables and the choice of causal mechanism. It is not always straightforward to decide what the "right" causal model is in a given situation, nor is it always obvious which of two causal models is "better" in some sense. These may be difficult decisions and often lie at the heart of determining actual causality in the real world. Nevertheless, we believe that the tools we provide here should help in making principled decisions about those choices.

**Syntax and Semantics:** To make the definition of actual causality precise, it is helpful to have a logic with a formal syntax. Given a signature $S = (\mathcal{U}, \mathcal{V}, \mathcal{R})$, a formula of the form $X = x$, for $X \in \mathcal{V}$ and $x \in \mathcal{R}(X)$, is called a *primitive event*. A *basic causal formula (over S)* is one of the form $[Y_1 \leftarrow y_1, \ldots, Y_k \leftarrow y_k]\varphi$ where $\varphi$ is a Boolean combination of primitive events, $Y_1, \ldots, Y_k, X$ are variables in $\mathcal{V}$, with $Y_1, \ldots, Y_k$ are distinct, $x \in \mathcal{R}(X)$, and $y_i \in \mathcal{R}(Y_i)$. Such a formula is abbreviated as $[\vec{Y} \leftarrow \vec{y}]\varphi$. The special case where $k = 0$ is abbreviated as $\varphi$. Intuitively, $[Y_1 \leftarrow y_1, \ldots, Y_k \leftarrow y_k]\varphi$ says that $\varphi$ holds in the counterfactual world that would arise if $Y_i$ is set to $y_i$, $i = 1, \ldots, k$. A *causal formula* is a Boolean combination of basic causal formulas.

A causal formula $\psi$ is true or false in a causal model, given a context. We write $(M, \vec{u}) \models \psi$ if $\psi$ is true in causal model $M$ given context $\vec{u}$. $(M, \vec{u}) \models [\vec{Y} \leftarrow \vec{y}](X = x)$ if the variable $X$ has value $x$ in the (unique, since we are dealing with recursive models) solution to the equations in $M_{\vec{Y} \leftarrow \vec{y}}$ in context $\vec{u}$ (that is, the unique vector of values for the exogenous variables that simultaneously satisfies all equations $F_Z^{\vec{Y} \leftarrow \vec{y}}$, $Z \in \mathcal{V} - \vec{Y}$, with the variables in $\mathcal{U}$ set to $\vec{u}$). $(M, \vec{u}) \models [\vec{Y} \leftarrow \vec{y}]\varphi$ for an arbitrary Boolean combination $\varphi$ of formulas of the form $\vec{X} = \vec{x}$ is defined similarly. We extend the definition to arbitrary causal formulas, i.e., Boolean combinations of basic causal formulas, in the standard way.

Note that the structural equations are deterministic. We can make sense out of probabilistic counterfactual statements, even conditional ones (the probability that $X$ would be 3 if $Y_1$ were 2, given that $Y$ is in fact 1) in this framework (see [Balke and Pearl 1994]), by putting a probability on the set of possible contexts. This is not necessary for our discussion of causality, although it plays a more significant role in the discussion of explanation.

## 3    THE DEFINITION OF CAUSE

With all this notation in hand, we can now give our definition of actual cause ("cause" for short). We want to make sense out of statements of the form "event $A$ is an actual cause of event $\varphi$ (in context $\vec{u}$)". As we said earlier, the context is the background information. While this has been left implicit in some treatments of causality, we find it useful to make it explicit. The picture here is that the context (and the structural equations) are given. Intuitively, they encode the background knowledge. All the relevant events are known. The only question is picking out which of them are the causes of $\varphi$ or, alternatively, testing whether a given set of events can be considered the cause of $\varphi$.

The types of events that we allow as actual causes are ones of the form $X_1 = x_1 \wedge \ldots \wedge X_k = x_k$—that is, conjunctions of primitive events; we typically abbreviate this as $\vec{X} = \vec{x}$. The events that can be caused are arbitrary Boolean combinations of primitive events.

**Definition 3.1:** (Actual cause) $\vec{X} = \vec{x}$ is an *actual cause of* $\varphi$ *in* $(M, \vec{u})$ if the following three conditions hold:

AC1. $(M, \vec{u}) \models (\vec{X} = \vec{x}) \wedge \varphi$. (That is, both $\vec{X} = \vec{x}$ and $\varphi$ are true in the actual world.)

AC2. There exists a partition $(\vec{Z}, \vec{W})$ of $\mathcal{V}$ with $\vec{X} \subseteq \vec{Z}$ and some setting $(\vec{x}', \vec{w}')$ of the variables in $(\vec{X}, \vec{W})$ such that if $(M, \vec{u}) \models Z = z^*$ for $Z \in \vec{Z}$, then

   (a) $(M, \vec{u}) \models [\vec{X} \leftarrow \vec{x}', \vec{W} \leftarrow \vec{w}']\neg\varphi$. In words, changing $(\vec{X}, \vec{W})$ from $(\vec{x}, \vec{w})$ to $(\vec{x}', \vec{w}')$ changes $\varphi$ from true to false,

   (b) $(M, \vec{u}) \models [\vec{X} \leftarrow \vec{x}, \vec{W} \leftarrow \vec{w}', \vec{Z}' \leftarrow \vec{z}^*]\varphi$ for all subsets $\vec{Z}'$ of $\vec{Z}$. In words, setting $\vec{W}$ to $\vec{w}'$ should have no effect on $\varphi$ as long as $\vec{X}$ is kept at its current value $\vec{x}$, even if all the variables in an arbitrary subset of $\vec{Z}$ are set to their original values in the context $\vec{u}$.

AC3. $\vec{X}$ is minimal; no subset of $\vec{X}$ satisfies conditions AC1 and AC2. Minimality ensures that only those elements of the conjunction $\vec{X} = \vec{x}$ that are essential for changing $\varphi$ in AC2(a) are considered part of a cause; inessential elements are pruned. ∎

Note that we allow $X = x$ to be a cause of itself. While we do not find such trivial causality terribly bothersome, it can be avoided by requiring that $\vec{X} = \vec{x} \wedge \neg\varphi$ be consistent for $\vec{X} = \vec{x}$ to be a cause of $\varphi$.

The core of this definition lies in AC2. Informally, the variables in $\vec{Z}$ should be thought of as describing the "active causal process" from $\vec{X}$ to $\varphi$. (also called "intrinsic process" by Lewis [1986]).[1] These are the variables that mediate between $\vec{X}$ and $\varphi$. Indeed, we can define an *active causal process* from $\vec{X} = \vec{x}$ to $\varphi$ as a minimal set $\vec{Z}$ that satisfies AC2. AC2(a) is reminiscent of the traditional counterfactual criterion of Lewis [1986], according to which $\varphi$ should be false if it were not for $\vec{X}$ being $\vec{x}$. However, AC2(a) is more permissive than the traditional criterion; it allows the dependence of $\varphi$ on $\vec{X}$ to be tested under special circumstances in which the variables $\vec{W}$ are held constant at some setting $\vec{w}'$. This modification of the traditional criterion was proposed by Pearl [1998, 2000] and was named *structural contingency*—an alteration of the model $M$ that involves the breakdown of some mechanisms (possibly emerging from external action) but no change in the context $\vec{u}$. The need to invoke such contingencies will be made clear in Example 3.2.

AC2(b), which has no obvious analogue in the literature, is an attempt to counteract the "permissiveness" of AC2(a) with regard to structural contingencies. Essentially, it ensures that $\vec{X}$ alone suffices to bring about the change from $\varphi$ to $\neg\varphi$; setting $\vec{W}$ to $\vec{w}'$ merely eliminates spurious side effects that tend to mask the action of $\vec{X}$. It captures the fact that setting $\vec{W}$ to $\vec{w}'$ does not affect the causal process by requiring that changing $\vec{W}$ from $\vec{w}$ to $\vec{w}'$ has no effect on the value of $\varphi$. Moreover, although the values in the variables $\vec{Z}$ involved in the causal process may be perturbed by the change, the perturbation has no impact on the value of $\varphi$. The upshot of this requirement is that we are not at liberty to conduct the counterfactual test of AC2(a) under an arbitrary alteration of the model. The alteration considered must not affect the causal process. Clearly, if the contingencies considered are limited to "freezing" variables at their actual value, so that $(M, \vec{u}) \models \vec{W} = \vec{w}'$, then AC2(b) is satisfied automatically. However, as the examples below show, genuine causation may sometimes be revealed only through a broader class of counterfactual tests in which variables in $\vec{W}$ are set to values that differ from their actual values. In [Pearl 2000], a notion of *contributory cause* is defined as well as actual cause. Roughly speaking, if AC2(a) holds only with $\vec{W} = \vec{w}' \neq \vec{w}$, the $A$ is a contributory cause of $B$; actual causality holds only if $\vec{W} = \vec{w}$.

We remark that, like the definition here, the causal beam definition [Pearl 2000] tests for the existence of counterfactual dependency in an auxiliary model of the world, modified by a select set of structural contingencies. However, whereas the beam criterion selects the choice of contingencies depends only on the relationship a variable and its parents in the causal diagram, the current definition selects the modifying contingencies based on the specific cause and effect pair that is being tested. This refinement permits our definition to avoid certain pitfalls (see Example 4.4) that are associated with graphical criteria for actual causation.

AC3 is a minimality condition. Heckerman and Shachter [1995] have a similar minimality requirement; Lewis [2000] mentions the need for minimality as well. Interestingly, in all the examples we have considered, AC3 forces the cause to be a single conjunct of the form $X = x$. Eiter and Lukasiewicz [2001] and, independently, Hopkins [2001], have recently proved that this is in fact a consequence of our definition.

How reasonable are these requirements? In particular, is it appropriate to invoke structural changes in the definition of actual causation? The following example may help illustrate why we believe it is.

**Example 3.2:** Suppose that two arsonists drop lit matches in different parts of a dry forest, and both cause trees to start burning. Consider two scenarios. In the first, called "disjunctive," either match by itself suffices to burn down the whole forest. That is, even if only one match were lit, the forest would burn down. In the second scenario, called "conjunctive," both matches are necessary to burn down the forest; if only one match were lit, the fire would die down. We can describe the essential structure of these two scenarios using a causal model with four variables:

- an exogenous variable $U$ which determines, among other things, the motivation and state of mind of the arsonists. For simplicity, assume that $\mathcal{R}(U) = \{u_{00}, u_{10}, u_{01}, u_{11}\}$; if $U = u_{ij}$, then the first arsonist intends to start a fire iff $i = 1$ and the second arsonist intends to start a fire iff $j = 1$. In both scenarios $U = u_{11}$.

- endogenous variables $ML_1$ and $ML_2$, each either 0 or 1, where $ML_i = 0$ if arsonist $i$ doesn't drop the match and $ML_i = 1$ if he does, for $i = 1, 2$.

- an endogenous variable $FB$ for forest burns down, with values 0 (it doesn't) and 1 (it does).

Both scenarios have the same causal network (see Figure 1); they differ only in the equation for $FB$. For the disjunctive scenario we have $F_{FB}(u, 1, 1) = F_{FB}(u, 0, 1) = F_{FB}(u, 1, 0) = 1$ and $F_{FB}(u, 0, 0) = 0$ (where $u \in \mathcal{R}(U)$); for the conjunctive scenario we have $F_{FB}(u, 1, 1) = 1$ and $F_{FB}(u, 0, 0) = F_{FB}(u, 1, 0) = F_{FB}(u, 0, 1) = 0$. In general, we expect that the causal model for reasoning about forest fires would involve many other variables; in particular, variables for other potential causes of forest fires such lightning and unattended campfires; here we focus on that part of the causal model that involves forest fires started by arsonists. Since for causality we assume that all the relevant facts are given, we can assume here that it is known that there were no unattended campfires and there was no lightning, which makes it safe to ignore that portion of the causal model. Denote by $M_1$ and $M_2$ the causal models associated with the disjunctive and conjunctive scenarios, respectively. The causal network for the relevant portion of $M_1$ and $M_2$ is described in Figure 1.

---

[1] Recently, Lewis [2000] has abandoned attempts to define "intrinsic process" formally. Pearl's "causal beam" [Pearl 2000, p. 318] is a special kind of active causal process, in which AC2(b) is expected to hold (with $\vec{Z} = \vec{z}^*$) for all settings $w'$ of $W$, not necessarily the one used in (a).

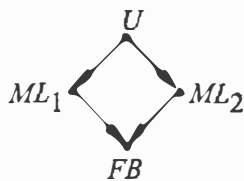

Figure 1: The causal network for $M_1$ and $M_2$.

Despite the differences in the underlying models, it is not hard to show that each of $ML_1 = 1$ and $ML_2 = 1$ is a cause of $FB = 1$ in both scenarios. We present the argument for $ML_1 = 1$ here. To show that $ML_1 = 1$ is a cause in $M_1$ let $\vec{Z} = \{ML_1, FB\}$, so $\vec{W} = \{ML_2\}$. It is easy to see that the contingency $ML_2 = 0$ satisfies the two conditions in AC2. AC2(a) is satisfied because, in the absence of the second arsonist ($ML_2 = 0$), the first arsonist is necessary and sufficient for the fire to occur ($FB = 1$). AC2(b) is satisfied because, if the first match is lit ($ML_1 = 1$) the contingency $ML_2 = 0$ does not prevent the fire from burning the forest. Thus, $ML_1 = 1$ is a cause of $FB = 1$ in $M_1$. (Note that we needed to set $ML_2$ to 0, contrary to facts, in order to reveal the latent dependence of $FB$ on $ML_1$. Such a setting constitutes a structural change in the original model, since it involves the removal of some structural equations.) A similar argument shows that $ML_1 = 1$ is also a cause of $FB = 1$ in $M_2$. (Again, taking $\vec{Z} = \{ML_1, FB\}$ and $\vec{W} = \{ML_2\}$ works.)

This example also illustrates the need for the minimality condition AC3. If lighting a match qualifies as the cause of fire then lighting a match and sneezing would also pass the tests of AC1 and AC2 and awkwardly qualify as the cause of fire. Minimality serves here to strip "sneezing" and other irrelevant, over-specific details from the cause. ∎

This is a good place to illustrate the need for structural contingencies in the analysis of actual causation. The reason we consider $ML_1 = 1$ to be a cause of $FB = 1$ in $M_1$ is that if $ML_2$ had been 0, rather than 1, $FB$ would depend on $ML_1$. In words, we imagine a situation in which the second match is not lit, and we then reason counterfactually that the forest would not have burned down if it were not for the first match.

## 4 EXAMPLES

In this section we show how our definition of actual causality handles some examples that have caused problems for other definitions. The full paper has further examples.

**Example 4.1:** The first example is due to Bennett (and appears in [Sosa and Tooley 1993, pp. 222–223]). Suppose that there was a heavy rain in April and electrical storms in the following two months; and in June the lightning took hold. If it hadn't been for the heavy rain in April, the forest would have caught fire in May. The question is whether the April rains caused *the* forest fire. According to a naive counterfactual analysis, they do, since if it hadn't rained, there wouldn't have been a forest fire in June. Bennett says "That is unacceptable. A good enough story of events and of causation might give us reason to accept some things that seem intuitively to be false, but no theory should persuade us that delaying a forest's burning for a month (or indeed a minute) is causing a forest fire."

In our framework, as we now show, it is indeed false to say that the April rains caused *the* fire, but they were a cause of there being a fire in June, as opposed to May. This seems to us intuitively right. To capture the situation, it suffices to use a simple model with three endogenous random variables:

- $AS$ for "April showers", with two values—0 standing for did *not* rain heavily in April and 1 standing for rained heavily in April;

- $ES$ for "electric storms", with four possible values: $(0,0)$ (no electric storms in either May or June), $(1,0)$ (electric storms in May but not June), $(0,1)$ (storms in June but not May), and $(1,1)$ (storms in April and May);

- and $F$ for "fire", with three possible values: 0 (no fire at all), 1 (fire in May), or 2 (fire in June).

We do not describe the context explicitly, either here or in the other examples. Assume its value $\vec{u}$ is such that it ensures that there is a shower in April, there are electric storms in both May and June, there is sufficient oxygen, there are no other potential causes of fire (like dropped matches), no other inhibitors of fire (alert campers setting up a bucket brigade), and so on. That is, we choose $\vec{u}$ so as to allow us to focus on the issue at hand and to ensure that the right things happened (there was both fire and rain).

We will not bother writing out the details of the structural equations—they should be obvious, given the story (at least, for the context $\vec{u}$); this is also the case for all the other examples in this section. The causal network is simple: there are edges from $AS$ to $F$ and from $ES$ to $F$. It is easy to check that each of the following hold.

- $AS = 1$ is a cause of the June fire ($F = 2$) (taking $\vec{W} = \{ES\}$ and $\vec{Z} = \{AS, F\}$) but not of fire ($F = 2 \lor F = 1$).

- $ES = (1,1)$ is a cause of both $F = 2$ and ($F = 1 \lor F = 2$). Having electric storms in both May and June caused there to be a fire.

- $AS = 1 \land ES = (1,1)$ is not a cause of $F = 2$, because it violates the minimality requirement of AC3; each conjunct alone is a cause of $F = 2$. Similarly, $AS = 1 \land ES = (1,1)$ is not a cause of ($F = 1 \lor F = 2$).

The distinction between April showers being a cause of the fire (which they are not, according to our analysis) and April showers being a cause of a fire in June (which they are) is one that seems not to have been made in the discussion of this problem (cf. [Lewis 2000]); nevertheless, it seems to us an important distinction. ∎

Although we did not describe the context explicitly in Example 4.1, it still played a crucial role. If the presence of oxygen is relevant then we must take this factor out of the context and introduce it as an explicit endogenous variables. Doing so can affect the causality picture. The next example already shows the importance of choosing an appropriate granularity in modeling the causal process and its structure.

**Example 4.2:** The following story from [Hall 2002] is an example of *preemption*, where there are two potential causes of an event, one of which preempts the other. An adequate definition of causality must deal with preemption in all of its guises.

> Suzy and Billy both pick up rocks and throw them at a bottle. Suzy's rock gets there first, shattering the bottle. Since both throws are perfectly accurate, Billy's would have shattered the bottle had it not been preempted by Suzy's throw.

Common sense suggests that Suzy's throw is the cause of the shattering, but Billy's is not. This holds in our framework too, but only if we model the story appropriately. Consider first a coarse causal model, with three endogenous variables:

- $ST$ for "Suzy throws", with values 0 (Suzy does not throw) and 1 (she does);

- $BT$ for "Billy throws", with values 0 (he doesn't) and 1 (he does);

- $BS$ for "bottle shatters', with values 0 (it doesn't shatter) and 1 (it does).

Again, we have a simple causal network, with edges from both $ST$ and $BT$ to $BS$. In this simple causal network, $BT$ and $ST$ play absolutely symmetric roles, with $BS = ST \vee BT$, and there is nothing to distinguish one from the other. Not surprisingly, both Billy's throw and Suzy's throw are classified as causes of the bottle shattering.

The trouble with this model is that it cannot distinguish the case where both rocks hit the bottle simultaneously (in which case it would be reasonable to say that both $ST = 1$ and $BT = 1$ are causes of $BS = 1$) from the case where Suzy's rock hits first. The model has to be refined to express this distinction. One way is to invoke a dynamic model [Pearl 2000, p. 326]. Another way to gain expressiveness is to allow $BS$ to be three valued, with values 0 (the bottle doesn't shatter), 1 (it shatters as a result of being hit by Suzy's rock), and 2 (it shatters as a result of being hit by Billy's rock). We leave it to the reader to check that $ST = 1$ is a cause of $BS = 1$, but $BT = 1$ is not (if Suzy hadn't thrown but Billy had, then we would have $BS = 2$). Thus, to some extent, this solves our problem. But it borders on cheating; the answer is almost programmed into the model by invoking the relation "as a result of", which requires the identification of the actual cause.

A more useful choice is to add two new random variables to the model:

- $BH$ for "Billy's rock hits the (intact) bottle", with values 0 (it doesn't) and 1 (it does); and

- $SH$ for "Suzy's rock hits the bottle", again with values 0 and 1.

With this addition, we can go back to $BS$ being two-valued. In this model, we have the causal network shown in Figure 2, with the arrow $SH \to BH$ being inhibitory; $BH = BT \wedge \neg SH$ (that is, $BH = 1$ iff $BT = 1$ and $SH = 0$). Note that, to simplify the presentation, we have omitted the exogenous variables from the causal network in Figure 2. In addition, we have only given the arrows for the particular context of interest, where Suzy throws first. In a context where Billy throws first, the arrow would go from $BH$ to $SH$ rather than going from $SH$ to $BH$, as it does in the figure.

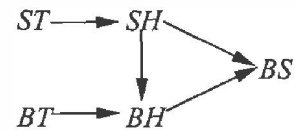

Figure 2: The rock-throwing example.

Now it is the case that $ST = 1$ is a cause of $BS = 1$. To satisfy AC2, we choose $\vec{W} = \{BT\}$ and $w' = 0$ and note that, because $BT$ is *set* to 0, $BS$ will track the setting of $ST$. Also note that $BT = 1$ is not a cause of $BS = 1$; there is no partition $\vec{Z} \cup \vec{W}$ that satisfies AC2. Attempting the symmetric choice $\vec{W} = \{BT\}$ and $w' = 0$ would violate AC2(b) (with $\vec{Z}' = \{BH\}$), because $\varphi$ becomes false when we set $ST = 0$ and restore $BH$ to its current value of 0.

This example illustrates the need for invoking subsets of $\vec{Z}$ in AC2(b). ∎

**Example 4.3:** Is causality transitive? Consider the following story, again taken from (an early version of) [Hall 2002]:

> Billy contracts a serious but nonfatal disease. He is hospitalized and treated on Monday, so is fine Tuesdsay morning. Had Monday's doctor forgotten to treat Billy, Tuesday's doctor would have treated him, and he would have been fine Tuesday afternoon. But there is a twist: one dose of medication is harmless, but two doses are lethal.

The causal model for this story is straightforward. There are three random variables: $MT$ for Monday's treatment (1 if Billy was treated Monday; 0 otherwise), $TT$ for Tuesday's treatment (1 if Billy was treated Tuesday; 0 otherwise), and $BMC$ for Billy's medical condition (0 if Billy is fine both Tuesday morning and Tuesday afternoon; 1 if Billy is sick

Tuesday morning, fine Tuesday afternoon; 2 if Billy is sick both Tuesday morning and afternoon; 3 if Bill is fine Tuesday morning and dead Tuesday afternoon). In the causal network corresponding to this causal model, shown in Figure 3, there is an edge from $MT$ to $TT$, since whether the Tuesday treatment occurs depends on whether the Monday treatment occurs, and edges from both $MT$ and $TT$ to $BMC$, since Billy's medical condition depends on both treatments.

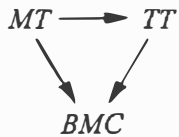

Figure 3: Billy's medical condition.

In this causal model, it is true that $MT = 1$ is a cause of $BMC = 0$, as we would expect—because Billy is treated Monday, he is not treated on Tuesday morning, and thus recovers Tuesday afternoon.[2] $MT = 1$ is also a cause of $TT = 0$, as we would expect, and $TT = 0$ is a cause of Billy's being alive ($BMC = 0 \lor BMC = 1 \lor BMC = 2$). However, $MT = 1$ is *not* a cause of Billy's being alive. It fails condition AC2(a): setting $MT = 0$ still leads to Billy's being alive (with $\vec{W} = \emptyset$). Note that it would not help to take $\vec{W} = \{TT\}$. For if $TT = 0$, then Billy is alive no matter what $MT$ is, while if $TT = 1$, then Billy is dead when $MT$ has its original value, so AC2(b) is violated (with $\vec{Z}' = \emptyset$).

This shows that causality is not transitive, according to our definitions. Although $MT = 1$ is a cause of $TT = 0$ and $TT = 0$ is a cause of $BMC = 0 \lor BMC = 1 \lor BMC = 2$, $MT = 1$ is not a cause of $BMC = 0 \lor BMC = 1 \lor BMC = 2$. Nor is causality closed under *right weakening*: $MT = 1$ is a cause of $BMC = 0$, which logically implies $BMC = 0 \lor BMC = 1 \lor BMC = 2$, which is not caused by $MT = 1$.

Lewis [1986, 2000] insists that causality is transitive, partly to be able to deal with preemption [Lewis 1986]. Our approach handles preemption without needing to invoke transitivity, which, as Lewis's own examples show, leads to counterintuitive conclusions. ∎

Clause AC2(b) in the definition is complicated by the need to check that no change in the value of the variables in $\vec{Z}$ can affect the value of $\varphi$. In all the previous examples, $Z = z^*$ for each variable $Z \in \vec{Z}$. Could we not just require this? The following example shows that we cannot.

**Example 4.4:** Imagine that a vote takes place. For simplicity, two people vote. The measure is passed if at least one of them votes in favor. In fact, both of them vote in favor, and the measure passes. This version of the story is almost identical to Example 3.2. If we use $V_1$ and $V_2$ to denote how the voters vote ($V_i = 0$ if voter $i$ votes against and $V_i = 1$ if she votes in favor) and $P$ to denote whether the measure passes ($P = 1$ if it passes, $P = 0$ if it doesn't), then in the context where $V_1 = V_2 = 1$, it is easy to see that each of $V_1 = 1$ and $V_2 = 1$ is a cause of $P = 1$. However, suppose we now assume that there is a voting machine that tabulates the votes. Let $M$ represent the total number of votes recorded by the machine. Clearly $M = V_1 + V_2$ and $P = 1$ iff $M \geq 1$. The following causal network represents this more refined version of the story. In this more refined scenario, $V_1 = 1$ and $V_2 = 1$ are still both causes of $P = 1$. Consider $V_1 = 1$. Take $\vec{Z} = \{V_1, M, P\}$ and $\vec{W} = V_2$. Much like the simpler version of the story, if we choose the contingency $V_2 = 0$, then $P$ is counterfactually dependent on $V_1$, so AC2(a) holds. To check if this contingency satisfies AC2(b), we set $V_1$ to 1 (their original value) and check that setting $V_2$ to 0 does not change the value of $P$. This is indeed the case. Although $M$ becomes 1, not 2 as it is when $V_1 = V_2 = 1$, nevertheless, $P = 1$ continues to hold, so AC2(b) is satisfied and $V_1 = 1$ is a cause of $P = 1$. However, if we had insisted in AC2(b) that $(M, u) \models [\vec{X} \leftarrow \vec{x}, \vec{W} \leftarrow w']Z = z^*$ for all variables $Z \in \vec{Z}$ (which in this case means that $M$ would have to retain its original value of 2 when $V_1 = 1$ and $V_2 = 0$), then neither $V_1 = 1$ nor $V_2 = 1$ would be a cause of $P = 1$. ∎

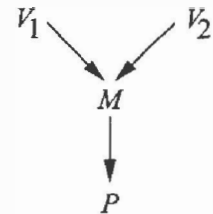

Figure 4: An example showing the need for AC2(b).

We remark that this example is not handled correctly by Pearl's causal beam definition. According to the causal beam definition, there is no cause for $P = 1$! It can be shown if $X = x$ is an actual (or contributory) cause of $Y = y$ according to the causal beam definition given in [Pearl 2000], then it is an actual cause according to the definition here. As Example 4.4 shows, the converse is not necessarily true.

**Example 4.5:** This example concerns what Hall calls the distinction between causation and determination. Again, we quote Hall [2002]:

> You are standing at a switch in the railroad tracks. Here comes the train: If you flip the switch, you'll send the train down the left-hand track; if you leave it where it is, the train will follow the right-hand track. Either way, the train will arrive at the same point, since the tracks reconverge up ahead. Your action is not among the causes of this arrival; it merely helps to determine how the arrival is brought about (namely, *via* the left-hand track, or *via* the right-hand track).

Again, our causal model gets this right. Suppose we have three random variables:

---
[2] Lewis's [1986] revised criterion of counterfactual-dependence-chain fails in this example; $BMC$ does not depend on either $MT$ or $TT$ in the context given.

- $F$ for "flip", with values 0 (you don't flip the switch) and 1 (you do);

- $T$ for "track", with values 0 (the train goes on the left-hand track) and 1 (it goes on the right-hand track);

- $A$ for "arrival", with values 0 (the train does not arrive at the point of reconvergence) and 1 (it does).

Now it is easy to see that flipping the switch ($F = 1$) does cause the train to go down the left-hand track ($T = 0$), but does not cause it to arrive ($A = 1$), thanks to AC2(a)—whether or not the switch is flipped, the train arrives.

However, our proposal goes one step beyond this simple picture. Suppose that we model the tracks using *two* variables:

- $LT$ for "left-track", with values 1 (the train goes on the left-hand track) and 0 (it does not); and

- $RT$ for "right-track", with values 1 (the train goes on the right-hand track) and 0 (it does not).

The resulting causal diagram is shown in Figure 5; it is isomorphic to a class of problems Pearl [2000] calls "switching causation". Lo and behold, this representation classifies $F = 1$ as a cause of $A$, which, at first sight, may seem counterintuitive: Can a change in representation turn a non-cause into a cause?

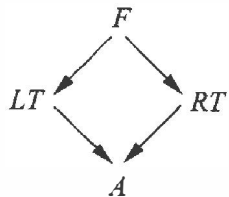

Figure 5: Flipping the switch.

It can and it should! The change to a two-variable model is not merely syntactic, but represents a profound change in the story. The two-variable model depicts the tracks as two independent mechanisms, thus allowing one track to be set (by action or mishap) to false (or true) without affecting the other. Specifically, this permits the disastrous mishap of flipping the switch while the left track is malfunctioning. Such abnormal eventualities are imaginable and expressible in the two-variable model, but not in the one-variable model. The potential for such eventualities is precisely what renders $F = 1$ a cause of the $A$ in the model of Figure 5.[3]

Is flipping the switch a legitimate cause of the train's arrival? Not in ideal situations, where all mechanisms work as specified. But this is not what causality (and causal modeling) are all about. Causal models earn their value in abnormal circumstances, created by structural contingencies, such as the possibility of a malfunctioning track. It is this possibility that should enter our mind whenever we decide to designate each track as a separate mechanism (i.e., equation) in the model and, keeping this contingency in mind, it should not be too odd to name the switch position a cause of the train arrival (or non-arrival). ∎

---

[3]This can be seen by noting that condition AC2 is satisfied by the partition $\vec{Z} = \{F, LT, A\}$, $\vec{W} = \{RT\}$, and choosing $w'$ as the setting $RT = 0$. The event $RT = 0$ conflicts with $F = 0$ under normal situations.

We conclude this section with an example that shows a potential problem for our definition, and suggest a solution.

**Example 4.6:** Fred has his finger severed by a machine at the factory ($FS = 1$). Fortunately, Fred is covered by a health plan. He is rushed to the hospital, where his finger is sewn back on. A month later, the finger is fully functional ($FF = 1$). In this story, we would not want to say that $FS = 1$ is a cause of $FF = 1$ and, indeed, according to our definition, it is not, since $FF = 1$ whether or not $FS = 1$ (in all contingencies satisfying AC2(b)).

However, suppose we introduce a new element to the story, representing a nonactual structural contingency: Larry the Loanshark may be waiting outside the factory with the intention of cutting off Fred's finger, as a warning to him to repay his loan quickly. Let $LL$ represent whether or not Larry is waiting and let $LC$ represent whether Larry cuts off Fred's finger. If Larry cuts off Fred's finger, he will throw it away, so Fred will not be able to get it sewn back on. In the actual situation, $LL = LC = 0$; Larry is not waiting and Larry does not cut off Fred's finger. So, intuitively, there seems to be no harm in adding this fanciful element to the story. Or is there? Suppose that, if Fred's finger is cut off in the factory, then Larry will not be able to cut off the finger himself (since Fred will be rushed off to the hospital). Now $FS = 1$ becomes a cause of $FF = 1$. For in the structural contingency where $LL = 1$, if $FS = 0$ then $FF = 0$ (Larry will cut off Fred's finger and throw it away, so it will not become functional again). Moreover, if $FS = 1$, then $LC = 0$ and $FF = 1$, just as in the actual situation.[4] ∎

This example seems somewhat disconcerting. Why should adding a fanciful scenario like Larry the Loanshark to the story affect (indeed, result in) the accident being a cause of the finger being functional one month later? While it is true that the accident would be judged a cause of Fred's good fortune by anyone who knew of Larry's vicious plan (many underworld figures owe their lives to "accidents" of this sort), the question remains how to distinguish genuine plans that just happened not to materialize from sheer fanciful scenarios that have no basis in reality. To some extent, the answer here is the same as the answer to essentially all the other concerns we have raised: it is a modeling issue. If we know of Larry's plan, or it seems like a reasonable possibility, we should add it to the model (in which case the accident is a cause of the finger being functional); otherwise we shouldn't.

But this answer makes the question of how reasonable a possibility Larry's plan are into an all-or-nothing decision. One solution to this problem is to extend our notion of causal model somewhat, so as to be able to capture more directly the intuition that the Larry the Loanshark scenario is indeed rather fanciful. There a number of ways of doing this; we choose one based on Spohn's notion of a *ranking function* (or *ordinal conditional function*) [Spohn 1988]. A *ranking* $\kappa$ on a space $W$ is a function mapping subsets of $W$ to $\mathbb{N}^* = \mathbb{N} \cup \{\infty\}$ such that $\kappa(W) = 0$, $\kappa(\emptyset) = \infty$, and $\kappa(A) = \min_{w \in A}(\kappa(\{w\}))$. Intuitively, an ordinal ranking assigns a degree of surprise to each subset of worlds in $W$, where 0 means unsurprising and higher numbers denote greater surprise. Let a *world* be a complete setting of the exogenous variables. Suppose that, for each context $\vec{u}$, we have a ranking $\kappa_{\vec{u}}$ on the set of worlds. The unique setting of the exogenous variables that holds in context $\vec{u}$ is given rank 0 by $\kappa_{\vec{u}}$; other worlds are assigned ranks according to how "fanciful" they are, given context $\vec{u}$. Presumably, in Example 4.6, an appropriate ranking $\kappa$ would give a world where Larry is waiting to cut off Fred's finger (i.e., where $LL = 1$) a rather high $\kappa$ ranking, to indicate that it is rather fanciful. We can then modify the definition of causality so that we can talk about $\vec{X} = \vec{x}$ being an actual cause of $\varphi$ in $(M, u)$ *at rank $k$*. The definition is a slight modification of condition AC2 in Definition 3.1 so the contingency $(\vec{x}', \vec{w}')$ must hold in a world of rank at most $k$; we omit the formal details here. We then can restrict actual causality so that the structural contingencies involved have at most a certain rank. This is one way of ignoring fanciful scenarios.

---

[4]We thank Eric Hiddleston for bringing this issue and this example to our attention.

## 5 DISCUSSION

We have presented a principled way of determining actual causes from causal knowledge. The essential principles of our account include using structural equations to model causal mechanisms; using uniform counterfactual notation to encode and distinguish facts, actions, outcomes, and contingencies; using structural contingencies to uncover causal dependencies; and careful screening of these contingencies to avoid tampering with the causal processes to be uncovered. While our definitions has some unsatisfying features (see Example 4.6), we hope that the examples presented here illustrate how well it deals with many of the problematic cases found in the literature. As the examples have shown, much depends on choosing the "right" set of variables with which to model a situation, which ones to make exogenous, and which to make endogenous. While the examples have suggested some heuristics for making appropriate choices, we do not have a general theory for how to make these choices. We view this as an important direction for future research.

### Acknowledgments

We thank Christopher Hitchcock for many useful comments on earlier versions of the paper, Zoltan Szabo for stimulating discussions, and Eric Hiddleston for pointing out Example 4.6. Halpern's work supported in part by NSF under grants IRI-96-25901 and IIS-0090145 and ONR under grant N000140010341. Pearl's work supported in part by grants from NSF, ONR, AFOSR, and MICRO.